\documentclass[runningheads]{llncs}

\usepackage[T1]{fontenc}

\usepackage{graphicx}

\usepackage{hyperref}
\usepackage{color}
\usepackage{amsmath}

\begin{document}

\author{Roseval Malaquias Junior\inst{1,2}\orcidID{0000-0002-6005-0515} \and
Ramon Pires\inst{2}\orcidID{0000-0002-0023-1971} \and
Roseli A. F. Romero\inst{1}\orcidID{0000-0001-9366-2780} \and
Rodrigo Nogueira\inst{2}\orcidID{0000-0002-2600-6035}
}

\authorrunning{R.\ Malaquias Jr.\ et al.}

\institute{Computer Science Department, University of São Paulo, São Carlos, São Paulo, Brazil \\
\email{roseval@usp.br}, \email{rafrance@icmc.usp.br} \and
Maritaca AI, Campinas, São Paulo, Brazil \\
\email{\{ramon, rodrigo\}@maritaca.ai}
}

\title{Juru: Legal Brazilian Large Language Model from Reputable Sources}

\maketitle
\begin{abstract}
    The high compute cost associated with pretraining large language models limits their research. Two strategies have emerged to address this issue: domain specialization and pretraining with high-quality data. To explore these strategies, we specialized the Mistral-7B model with 1.9 billion unique tokens from reputable Brazilian legal sources and conducted few-shot evaluations on legal and general knowledge test suites. Our model, Juru, demonstrates the benefits of domain specialization by achieving improved performance on legal benchmarks, even with a reduced amount of pretraining data. However, this domain specialization through continued pretraining comes at the cost of increased forgetting in unrelated domains, as evidenced by performance degradation on general knowledge test suites in both Portuguese and English. This study contributes to the growing body of scientific evidence showing that pretraining data selection may enhance the performance of large language models, enabling the exploration of these models at a lower cost. Juru is publicly available at \url{https://huggingface.co/roseval/Juru-7B}.
    
\end{abstract}

\section{Introduction}

Large Language Models (LLMs) are commonly trained on vast amounts of general-purpose data sourced from internet snapshots, such as the CommonCrawl \cite{brown2020language,274574,touvron2023llama} and C4 \cite{touvron2023llama,xue-etal-2021-mt5}. Scaling laws \cite{hoffmann2022training} support this paradigm, suggesting more capable models with increased data and trainable parameters. As a result, LLMs developed through this method demonstrate the ability to address tasks across multiple languages and domains \cite{openai2023gpt4}.

The significant computational resources required for pretraining with such massive datasets pose a challenge for research, limiting the exploration of their capabilities. However, it is possible to observe competitive results on a smaller scale through the application of continued pretraining. 
In the literature, the selection of domain-specific data~\cite{colombo2024saullm,garcia-etal-2024-robertalexpt,pires2023sabia} from reputable sources \cite{gunasekar2023textbooks,li2023textbooks,refinedweb} stands out. 

The large general-purpose pretraining of LLMs promotes general factual knowledge and language comprehension. Therefore, continued pretraining on the same scale and scope might not be necessary to improve LLMs' performance.  
By employing domain-specific data from reputable sources, we postulate that an LLM initially trained on general-purpose data can improve its performance in a specific domain with minimal additional data, although this may come at the expense of performance degradation in other domains.  

To test this hypothesis, we specialized the Mistral-7B \cite{jiang2023mistral} model for the Brazilian legal domain, using $1.9$ billion unique tokens sourced mainly from academic studies from reputable sources. The resulting model, Juru, was evaluated on multiple-choice test suites covering both Brazilian legal knowledge and general knowledge domains. Our objective is to assess the impact of incorporating exclusively domain-specific data during continued pretraining on the model's performance. 

Across the proposed pretraining, our model demonstrated enhanced performance on the legal test suite, while exhibiting a decrease in performance on the general knowledge test suites. In particular, a more pronounced decrease was observed on the English test suite compared to the Portuguese one, suggesting that previously acquired knowledge is better retained when it shares greater similarity with the target domain of specialization. These results contribute to the growing body of scientific evidence supporting the effectiveness of specialized pretraining for improving the performance of generalist models on specific domains. Such specialization may facilitate their research by reducing the compute cost associated with training. Additionally, the resulting model, Juru, is publicly released to support further research on domain specialization through continued pretraining. To the best of our knowledge, Juru is the first LLM pretrained for the Brazilian legal domain. 

\section{Related Work} 

The increasing scale of LLMs, in both parameter numbers and training data, has led to the emergence of previously unseen abilities in smaller-scale models \cite{wei2022emergent}. Recent advances have enabled these models to achieve human-level performance across various tasks that require textual comprehension, specialized knowledge, mathematical reasoning, multilingualism, programming, and image interpretation \cite{openai2023gpt4}. With such results, it is arguable that these models demonstrate the capacity to integrate these diverse skills in a cohesive manner. 

Language and domain specialization have become effective strategies for ensuring that LLMs bring tangible benefits in real-world applications. Recent works have shown that continued pretrained in language-specific corpora, particularly in Portuguese~\cite{almeida2024sabia2,larcher2023cabrita,pires2023sabia}, enhances significantly the performance of downstream tasks without the need to increase the parameter count. 

The Brazilian legal field emerges as a relevant example of such domain specialization, as previous studies have underscored the performance of language models pretrained on legal corpora~\cite{garcia-etal-2024-robertalexpt,polo2021legalnlp}. 
However, those models are encoder-only, lacking the ability to generate text sequentially, which is essential for several legal applications. 

SaulLM-7B \cite{colombo2024saullm} is the first LLM pretrained to tackle the nuances of legal languages and follow instructions to support conversation interactions. Starting from Mistral-7B \cite{jiang2023mistral}, the model was pretrained on 30 billion tokens of an English legal dataset. In a subsequent study, Colombo et al. \cite{colombo2024saullm54bsaullm141bscaling} scaled this approach to a Mixture-of-Experts architecture. The authors continually pretrained Mixtral-54B and Mixtral-141B on 540 billion tokens of English legal texts, producing SaulLM-54B and SaulLM-141B. While these larger models achieve higher performance on most in-domain tasks, their increased size correlates with lower performance on certain legal tasks.

To the best of our knowledge, Juru is the first LLM pretrained for the Brazilian legal domain. Our experiments show that, relative to its base model, Juru achieves superior performance on in-domain tasks. However, its performance deteriorates on tasks beyond the Brazilian legal domain.

\section{Methodology}

In this section, we describe the gathering and curation of data from public sources within the Brazilian legal domain, as well as the pretraining of the Juru model, including its base model, hyperparameters, and optimization methods. 

\subsection{Pretraining Data}

We conducted web scraping of academic papers in Portuguese within the Brazilian legal domain, prioritizing data with educational value. 

Academic papers were scraped from various reputable sources that catalog research from Brazilian higher education institutions and national journals. Each platform offers filters for extracting Brazilian papers exclusively in Portuguese, accessible for non-commercial distribution. Also, we scraped the LexML\footnote{\url{https://www.lexml.gov.br}} database, focusing on a subset of Brazilian federal laws. 

A total of 13,278 documents were extracted from academic paper databases. Nearly $99\%$ of these documents are in Portable Document Format (PDF). Hence, we used Marker\footnote{\url{https://github.com/VikParuchuri/marker}} for extracting text from these documents. 

Texts from 14,743 federal laws were extracted from the LexML database. Additionally, the dataset from the study by Sakiyama et al. \cite{sakiyama2023exploring} was added to the pretraining dataset, consisting of 32,535 legal documents distributed between decisions and judgments from the Brazilian Supreme Federal Court. 

After extracting the documents, they were automatically curated using the filter proposed by Rae et al.~\cite{rae2021scaling}, adapted for Portuguese. This filter removes documents that are unlikely to be predominantly natural language, such as tables or long lists, as they could hinder the learning process of the LLM.

\begin{table}[htb]
\centering
\caption{Distribution of Byte-Pair Encoding (BPE) \cite{sennrich2016neural} tokens for the pretraining dataset, using the Mistral-7B tokenizer.} 

\renewcommand{\arraystretch}{1.5}
\begin{tabular}{cc}
\hline Dataset & Tokens \\
\hline Academic papers & 1,805,059,922 \\
Federal regulations & 60,402,071 \\
Decisions and rulings & 133,295,895 \\
\hline Total & 1,998,757,888 \\
\hline
\end{tabular}
\label{tb:tokens}
\end{table}

Considering the tokenizer of Mistral-7B, the model employed in our pretraining, we gathered $1.9$ billion tokens, as shown in Table \ref{tb:tokens}. Each document was divided into sequences of 4,096 tokens. 

\subsection{Juru Model}

The Juru model extends the causal language modeling pretraining of a Mistral-7B model, without instruction finetuning. We employed the t5x and seqio frameworks \cite{roberts2023scaling} for the proposed pretraining. 

Although more recent models \cite{grattafiori2024llama3herdmodels,groeneveld-etal-2024-olmo} of comparable size have demonstrated superior performance compared to Mistral-7B, we chose Mistral-7B due to its earlier release date. This decision aims to reduce the risk of data contamination in our test suites.

For the pretraining hyperparameters, we employed the AdaFactor optimizer \cite{shazeer2018adafactor} without factorization, with a first-order momentum of $0.9$ and a second-order momentum of $1 - k^{-0.8}$, where $k$ represents the current training step. Additionally, we applied dynamic weight decay by $lr^2$, with $lr$ as the current learning rate value, and global norm clipping of $1.0$. 

In addition to the loss associated with the causal language modeling task, we used an auxiliary loss of $10^{-4} \log ^2\left(\sum_i e^{z_i}\right)$, where $z$ denotes the logits of the prediction. This auxiliary loss helps dampen the values of the pretraining loss. The learning rate is initially set at $0$ and gradually increases during a linear warm-up period of $250$ training steps until it reaches $0.001$. Following this warm-up phase, the learning rate remains constant. 

The model was pretrained for causal language modeling on a cluster of TPUs v2-256. Each batch consisted of 512 sequences, with 4,096 tokens in each sequence. Training was conducted for 3,800 steps, corresponding to approximately 4 epochs, processing a total of 7.96 billion tokens. The entire pretraining consumed $3.35 \times 10^{20}\,\text{FLOPs}$ of compute \cite{kaplan2020scalinglawsneurallanguage}, taking $30.61$ hours to complete. The training achieved a Model FLOPs Utilization (MFU) of $54.2\%$, excluding self-attention operations \cite{chowdhery2022palm}. 

\section{Evaluation}

Evaluating LLMs for text generation poses challenges due to the subjective nature of emerging skills such as natural language understanding and domain expertise, alongside their ability to integrate these skills for problem-solving. Given their broad capabilities, evaluating with open-ended questioning provides insights into the models' knowledge representation and their capacity to generate human-like responses \cite{NEURIPS2023_91f18a12}. 

However, the evaluation of responses to open-ended questions poses inherent difficulties. Human evaluation is resource-intensive, requiring expertise in specific domain tasks and training in evaluation methodology for experimental robustness. Additionally, relying solely on SOTA LLMs for evaluation may yield a low correlation with human judges due to specific vocabulary and knowledge of a target domain. Hence, there's an increasing emphasis on evaluating these models using standardized multiple-choice exams, such as those used in university admission processes \cite{almeida2023bluex,silveira2018advances,nunes2023evaluating,pires2023evaluating,sayama2019faquad}. Such exams cover diverse topics and skills, proving advantageous for LLM evaluation. 

We investigate how continued pretraining through domain specialization influences both knowledge learning and forgetting. Hence, we evaluate the specialized Juru model on three standardized multiple-choice test suites: Brazilian legal knowledge, representing the target domain of our specialization; Portuguese general knowledge, which share the same language as the target domain but do not assess legal knowledge; and English general knowledge, whose distribution is closer to the original English-centric training data of Mistral-7B. 

For all test suites, we adopt accuracy as the primary evaluation metric. We group all editions of the same exam administered within a given year into a single evaluation group. When aggregating multiple exams, we report the mean accuracy. Lastly, we employ few-shot learning with three examples during our evaluation, as the model was trained solely for text completion and did not undergo any instruction finetuning.

\subsection{Brazilian Legal Knowledge}

For evaluating the Juru model, we used questions from six Bar Association Exams (OAB) of $2024$ and $2023$, and two National Judiciary Exams (ENAM) of $2024$. The OAB exam is mandatory for all law graduates aspiring to practice law in Brazil. It comprises two phases; the initial phase encompasses $80$ multiple-choice questions, while the second phase consists of $4$ essay questions and the creation of a procedural document. ENAM was introduced in 2023 to unify the selection process for entry into the Brazilian judiciary. As of its implementation, only candidates who have passed the ENAM within the last two years are eligible to apply for judicial positions in Brazil. This exam includes a single phase with $80$ multiple-choice questions. We included only multiple-choice questions from the $2023$ and $2024$ OAB, and $2024$ ENAM exams, comprising a total set of $638$ questions, as illustrated in Table \ref{tb:legal_evaluation}.

\begin{table}[htb] 
\centering
\caption{Configuration of Brazilian legal knowledge test suite.}

\renewcommand{\arraystretch}{1.5}
\begin{tabular}{c c c c c}
\hline
Benchmark & Domain & Exams & Task & Size \\ \hline
OAB-2023 & Legal & 3& Multiple-choice (4) & 240 \\
OAB-2024 & Legal & 3& Multiple-choice (4) & 238 \\
ENAM-2024 & Legal & 2& Multiple-choice (5) & 160 \\ \hline
\end{tabular}
\label{tb:legal_evaluation}
\end{table} 

\subsection{Portuguese General Knowledge}

Pires et al. \cite{pires2023sabia} observed a performance degradation on English benchmarks following the specialization of an English-centric LLM for Portuguese. These results raise the following question: Is catastrophic forgetting limited to cross-linguistic domain specialization, or does specializing in a specific domain within a single language also lead to forgetting of general knowledge in that language? 

To address this question, we examined the impact of specializing the Juru model in the Brazilian legal domain on its performance in Portuguese general knowledge tasks, using the multiple-choice exams listed in Table \ref{tb:portuguese_general_evaluation}, comprising a total amount of 2,123 questions.

\begin{table}[htb] 
\centering
\caption{Configuration of Portuguese general knowledge test suite.}
%\vspace*{1mm}
\renewcommand{\arraystretch}{1.5}
\begin{tabular}{c c c c c}
\hline
Benchmark & Domain & Exams & Task & Size \\ \hline
ENEM-2024 & General & 1 & Multiple-choice (5) & 179 \\
BLUEX-2024 & General & 2 & Multiple-choice (5) & 172 \\
CPNU-2024 & General & 7 & Multiple-choice (5) & 320 \\
BNDES-2024 & General & 12 & Multiple-choice (5) & 420 \\
REVALIDA-2024 & Medical & 2 & Multiple-choice (4) & 187 \\
MREX-2024 & Medical & 6 & Multiple-choice (4) & 452 \\
CFCEQ-2024 & Accounting & 12 & Multiple-choice (5) & 294 \\
CFCES-2024 & Accounting & 2 & Multiple-choice (4) & 99 \\\hline
\end{tabular}
\label{tb:portuguese_general_evaluation}
\end{table}

Following Almeida et al. \cite{almeida2024sabia2}, we ignore all the questions that require image understanding. Some exams have captions describing the image contents. In those cases, we incorporate them into the exact position the images would appear. 

\subsection{English General Knowledge}

Unlike the Sabiá \cite{pires2023sabia} model, which underwent extensive continued pretraining, the Juru model was trained for fewer steps due to the smaller size of its legal specialization dataset. While Pires et al.~\cite{pires2023sabia} observed a loss of English knowledge in their model after extended training on Portuguese data, their experiments involved a significantly longer training regime and a more diverse dataset. This raises another question: Can shorter, domain-specific pretraining mitigate or prevent catastrophic forgetting of previously learned knowledge? 

To investigate this, we evaluated the Juru model on the MMLU benchmark~\cite{hendryckstest2021}, a widely used test suite for assessing general knowledge and reasoning abilities in English~\cite{jiang2023mistral,openai2023gpt4,touvron2023llama2}. While MMLU includes a broad range of subjects, such as STEM, humanities, law, and medicine, we restricted our evaluation to the college-level and high school-level subsets, amounting to 4,369 questions, as outlined in Table~\ref{tb:mmlu_general_evaluation}. 

\begin{table}[htb]
\centering
\caption{Configuration of English general knowledge test suite.}

\renewcommand{\arraystretch}{1.5}
\begin{tabular}{c c c c c}
\hline
Benchmark & Domain & Exams & Task & Size \\ \hline
MMLU-College & General & 6 & Multiple-choice (4) & 719 \\
MMLU-High School & General & 14 & Multiple-choice (4) & 3,650 \\
\hline
\end{tabular}
\label{tb:mmlu_general_evaluation}
\end{table}

\section{Results} 

\begin{figure}[htp!]
    \centering
    \includegraphics[width=\linewidth]{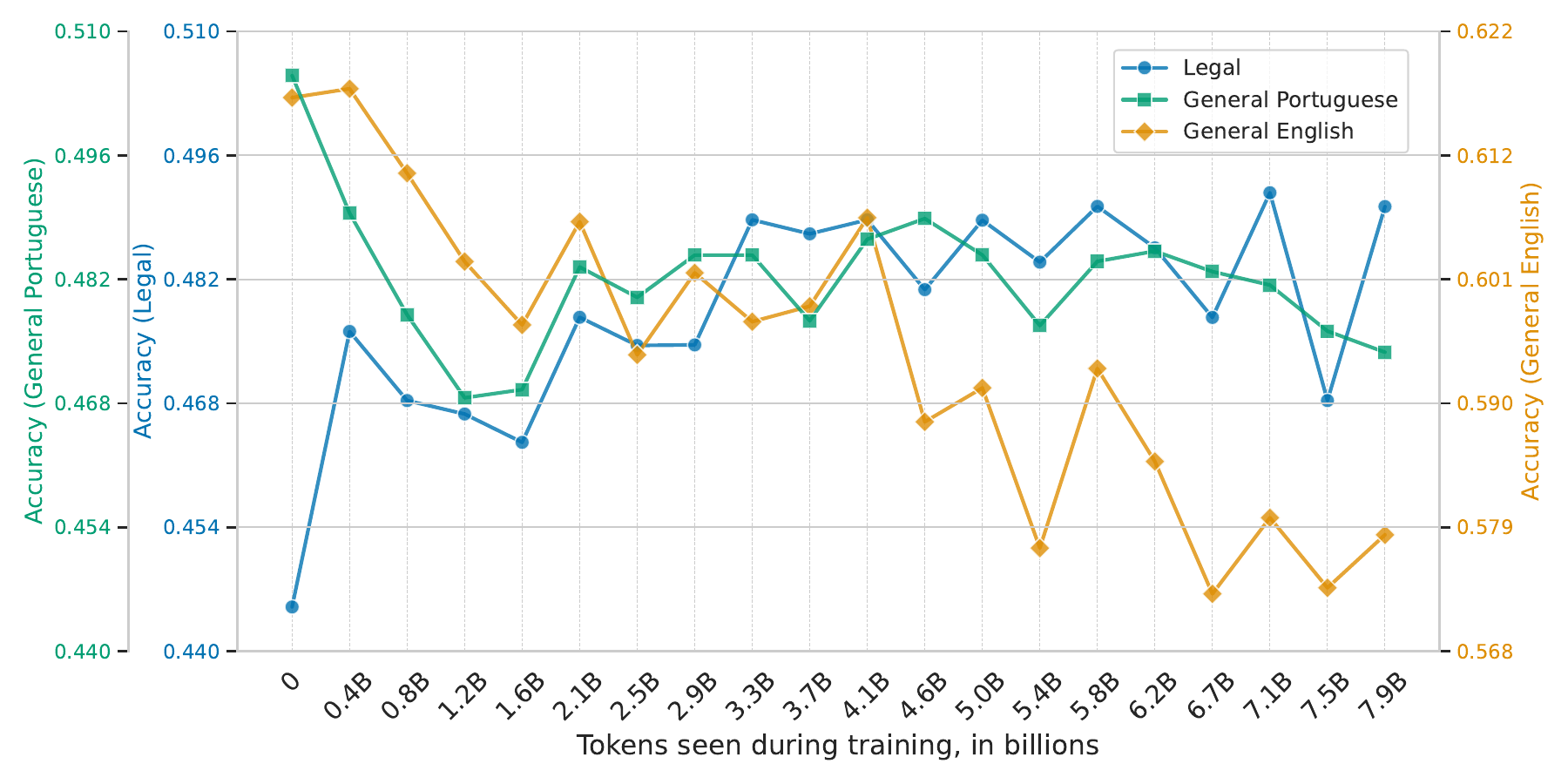}
    \caption{Accuracy on the Brazilian legal knowledge, English general knowledge, and Portuguese general knowledge test suites during continued pretraining.}
    \label{fig:juru}
\end{figure}

The proposed test suites were used to track the model’s performance across checkpoints during continued pretraining. Figure \ref{fig:juru} presents the accuracy as a function of the number of tokens processed during continued pretraining. Each epoch comprises $1.9$ billion tokens from Brazilian legal texts. The 0B point corresponds to the base Mistral-7B model, while the checkpoint at 7.1B tokens represents the specialized Juru model.

The results support the study’s central hypothesis: as the LLM undergoes domain specialization, performance improves on in-domain tasks but declines on out-of-domain general knowledge tasks. Upon analyzing the curves, we observed the model’s accuracy on the legal test suite steadily increased, peaking at $49.2\%$ on 7.1B tokens. In contrast, performance on general knowledge test suites decreased relative to the base model, with final accuracies of $48.1\%$ for the Portuguese knowledge test suite and $58.0\%$ for the English knowledge test suite at the same checkpoint.

Overall, the optimal version of Juru was obtained at 7.1B tokens, corresponding to the highest accuracy on the legal knowledge test suite. All subsequent analyses are based on this checkpoint, which we refer to as Juru. The next subsection compares the performance of this final model to the base Mistral-7B model across both legal and general knowledge test suites. 

\subsection{Results per Benchmark}

Table \ref{tb:direito} shows that Juru outperforms the base Mistral-7B model across all legal benchmarks, with an overall gain of approximately 4.7\% in average accuracy. The most notable improvement is observed in the ENAM-2024 benchmark, with a gain of 5.6\%. This result demonstrates the effectiveness of domain specialization through continued pretraining in enhancing performance on legal knowledge tasks. 

\begin{table}[htb] 
\centering
\caption{Comparison of Mistral-7B against Juru on Brazilian legal knowledge benchmarks.} 
\renewcommand{\arraystretch}{1.5}
\begin{tabular}{l c c}        
\multicolumn{1}{l}{}        & \multicolumn{2}{c}{Accuracy} \\ \hline
Benchmark  & Mistral-7B & Juru \\ \hline
OAB-2023 & 48.3\% & \textbf{54.5\%} \\  
OAB-2024 & 49.6\% & \textbf{52.0\%} \\  
ENAM-2024  & 31.2\% & \textbf{36.8\%} \\ \hline
Mean (8 Exams)& 44.5\%  & \textbf{49.2\%} \\ \hline
\end{tabular} 
\label{tb:direito}
\end{table}

In contrast, Table \ref{tb:english} indicates a decline in performance on the English general knowledge benchmarks after domain specialization. While Mistral-7B achieves higher scores on both the college and high school subsets of MMLU, Juru underperforms across the board, resulting in a 3.6\% decrease in the overall average accuracy. 

\begin{table}[htb] 
\centering
\caption{Comparison of Mistral-7B against Juru on English general knowledge benchmarks.} 
\renewcommand{\arraystretch}{1.5}
\begin{tabular}{l c c}        
\multicolumn{1}{l}{}        & \multicolumn{2}{c}{Accuracy} \\ \hline
Benchmark  & Mistral-7B & Juru \\ \hline
MMLU-College & \textbf{54.2\%} & 49.4\% \\  
MMLU-High School   & \textbf{64.8\%} & 61.6\% \\ \hline
Mean (20 Exams)& \textbf{61.6\%}  & 58.0\% \\ \hline
\end{tabular} 
\label{tb:english}
\end{table}

A similar trend is observed for the Portuguese general knowledge benchmarks in Table \ref{tb:portuguese}. Juru shows lower accuracy in 5 out of 8 individual benchmarks, leading to a drop of 2.4\% in the average accuracy across all exams. 

\begin{table}[htb] 
\centering
\caption{Comparison of Mistral-7B against Juru on Portuguese general knowledge benchmarks.} 
\renewcommand{\arraystretch}{1.5}
\begin{tabular}{l c c}        
\multicolumn{1}{l}{}        & \multicolumn{2}{c}{Accuracy} \\ \hline
Benchmark  & Mistral-7B & Juru \\ \hline
ENEM-2024 & \textbf{62.0\%} & 61.4\% \\  
BLUEX-2024  & \textbf{60.7\%} & 58.4\% \\  
CPNU-2024 & 56.8\% & 56.8\% \\  
BNDES-2024  & \textbf{59.1\%} & 56.0\% \\  
REVALIDA-2024 & 50.7\% & \textbf{51.7\%} \\  
MREX-2024  & \textbf{42.2\%} & 36.7\% \\  
CFCEQ-2024 & \textbf{40.7\%} & 38.0\% \\  
CFCES-2024 & 40.3\% & \textbf{41.3\%} \\ \hline 
Mean (44 Exams)& \textbf{50.5\%}  & 48.1\% \\ \hline
\end{tabular} 
\label{tb:portuguese}
\end{table}

\begin{figure}[!htb]
    \centering
    \includegraphics[width=\linewidth]{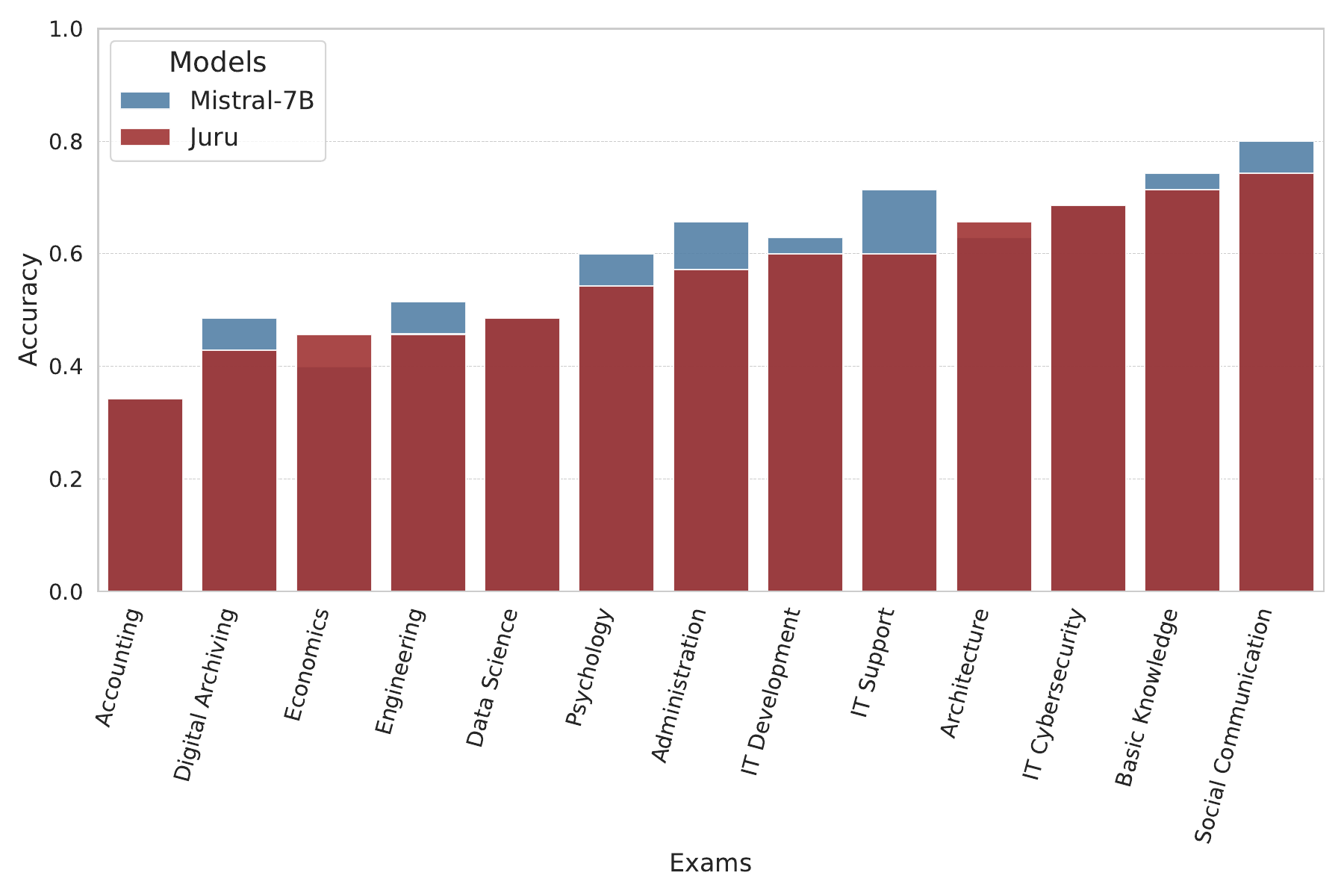}
    \caption{Comparison of Juru against Mistral-7B on BNDES-2024 exams.}
    \label{fig:bndes} 
\end{figure}

Figure \ref{fig:bndes} presents the comparison between Juru and Mistral-7B across 13 areas of knowledge from the BNDES-2024 benchmark. Overall, Juru underperforms the base model in 8 out of the 13 exams. The most significant decline is observed in the IT Support, with a drop of approximately 11.4\%. In contrast, Juru maintains the same accuracy as Mistral-7B in IT Cybersecurity, Data Science, and Accounting, and shows gains in Architecture and Economics. Despite these specific improvements, the overall trend highlights a decrease in performance across general knowledge areas following domain-specific continued pretraining.

\subsection{Results Analysis}

The results demonstrate the dynamics of learning domain-specific knowledge for the specialization of the Juru model. Specialization in the Brazilian legal domain yielded an increase of $4.7\%$ accuracy compared to Mistral-7B in the Brazilian legal knowledge test suite, despite the relatively small training dataset of only $1.9$ billion unique tokens. 

In line with the hypothesis of this study, specialization has led to a decline in the model's capacity to solve tasks beyond its specialized domain. Despite the shared language among evaluated tasks on the legal and Portuguese general test suites, specialization in only one knowledge domain has resulted in decreased performance in other domain areas. Clear correlations between types of knowledge areas and degradation were not evident, as shown in Figure \ref{fig:bndes}. Furthermore, when comparing the decline in performance across the Portuguese and English general knowledge test suites, the Juru model exhibited a greater degree of forgetting on the English suite. This pattern suggests that knowledge forgetting is influenced by the similarity between the evaluated and specialized domains: the less similar the previously learned knowledge is to the target domain, the greater the degree of forgetting. 

We posit that further pretraining of the Juru model is likely to increase the disparity between legal and general knowledge curves. Consequently, while performance in legal-related tasks may increase, there would likely be a significant decrease in performance in specific knowledge domains that require different skill sets. 

\section{Limitations} 

While positive results are evident, there's a possible risk of contamination in the pretraining dataset. This risk arises from web scraping pretraining data in the latter half of $2023$, after the publication of the OAB-2023 and MMLU exams. Although the dataset was primarily sourced from reputable Brazilian academic and legal documents, and it is unlikely to contain the test questions, the risk of indirect contamination cannot be fully ruled out. To address this, we plan to include newly released exams from 2025 and 2026 in future evaluations. 

All experiments reported in this paper are based on a specific subset of Brazilian legal data. Whether the observed trends generalize to other subdomains or other types of data than those originated from reputable sources remains an open question. Exploring a broader range of legal corpora, including news articles, blogs, and jurisprudential commentaries, may reveal different specialization behaviors.

Our work is subject to the same validity challenges commonly faced in the evaluation of LLMs, as noted in prior studies \cite{colombo2024saullm,grattafiori2024llama3herdmodels,groeneveld-etal-2024-olmo}. In the Brazilian context, what constitutes legal proficiency? Is it the ability to recall legislation, generate legally sound summaries, or predict judicial decisions? We argue that these skills are interrelated. Therefore, improving performance on one legal task may correlate with improvements in others, as they often rely on overlapping capabilities. We do not claim that Juru excels across all legal language processing tasks. Rather, we use performance on Brazilian legal standardized exams, designed to assess human legal competence, as a practical proxy for specialization effectiveness. The observed improvements on these benchmarks suggest that domain specialization may transfer to a broader range of legal tasks.

\section{Conclusions and Future Work}

In this study, we introduced Juru, the first LLM specialized in the Brazilian legal domain, pretrained on $1.9$ billion unique tokens. Despite its smaller size and limited data, we observed an expressive improvement in the model's ability to resolve multiple-choice questions within the Brazilian legal domain. Our findings confirm that domain specialization can be an efficient strategy for improving LLM performance at a lower compute cost. However, this comes with a trade-off: diminished performance on tasks unrelated to the target domain. 

Additionally, the comparison between Portuguese and English general test suites revealed a more pronounced decline in the English benchmarks. This suggests that the extent of forgetting may be linked to domain similarity: the lower the similarity the previously acquired knowledge has to the specialized domain, the higher the risk of forgetting.

For future work, our primary aim is to incorporate evaluations of the Juru model with new benchmarks that fall outside the model's knowledge cut-off to mitigate the possibility of data contamination. A promising direction is to investigate whether aggregating domains with high similarity during continued pretraining could enhance specialization performance without catastrophic forgetting of previously learned knowledge. Deeper understanding of these dynamics may lead to the development of LLMs with less compute through targeted domain specialization. We also release all Juru checkpoints from the experiments presented in this paper to further support future research on these topics, particularly within the Brazilian legal domain.  

\subsubsection{\ackname}
This work was partially financed by the Coordenação de Aperfeiçoamento de Pessoal de Nível Superior - Brasil (CAPES) and INCT (CAPES \#88887.136349/2017-00, CNPQ \#465755/2014-3). We thank Google Cloud Platform for the TPU grant.

\subsubsection{\discintname}
The authors declare no conflicts of interest related to the content of this article.

\bibliographystyle{splncs04}
\bibliography{references}

\end{document}